 \documentclass[smallabstract,smallcaptions]{dccpaper}

\usepackage{epsfig}
\usepackage{amsmath}
\usepackage{amssymb}
\usepackage{color}
\usepackage{url}

\usepackage{amssymb}
\usepackage{amsthm}
\usepackage{amsmath}
\usepackage{hyperref}
\usepackage{url}
\usepackage{graphicx}
\usepackage{subfigure}
\usepackage{algorithm}
\usepackage{algorithmic}
\usepackage{ulem}
\usepackage{array}
\usepackage{booktabs}
\usepackage{wrapfig}

\newlength{\figurewidth}
\newlength{\smallfigurewidth}

\begin{document}

\title
{\large
\textbf{
RQAT-INR: Improved Implicit Neural Image Compression }
}

\author{%
Bharath Bhushan Damodaran, Muhammet Balcilar, Franck Galpin, and Pierre Hellier\\[0.5em]
{\small\begin{minipage}{\linewidth}\begin{center}
\begin{tabular}{ccc}
InterDigital, Inc., \\
Rennes, France \\
\url{bharath.damodaran@interdigital.com} \\
\url{firstname.lastname@interdigital.com} 
\end{tabular}
\end{center}\end{minipage}}
}

\maketitle
\thispagestyle{empty}

\begin{abstract}
Deep variational autoencoders for image and video compression have gained significant attraction in the recent years, due to their potential to offer competitive or better compression rates compared to the decades long traditional codecs such as AVC, HEVC or VVC. However, because of complexity and energy consumption, these approaches are still far away from practical usage in industry. More recently, implicit neural representation (INR) based codecs have emerged, and have lower complexity and energy usage to classical approaches at decoding. However, their performances are not in par at the moment with state-of-the-art methods. In this research, we first show that INR based image codec has a lower complexity than VAE based approaches, then we propose several improvements for INR-based image codec and outperformed baseline model by a large margin. 
\end{abstract}

\section{Introduction}
Developing efficient lossy image and video compression is a long standing research problem, whose importance is still growing due to the increase in the terabytes of multimedia\cite{damodaran2021filmroll} contents generated every day. Their main aim is to remove the redundant information (e.g spatial and temporal redundancies and to encode the data with minimal bit-stream, at the same-time able to reconstruct the data with minimal distortion. Traditional compression techniques such as AVC, HEVC or VVC, relies on data independent transformation to remove the spatial redundancy, and they are able to offer competitive compression rates \cite{vvc}. Recently, learning based compression techniques based on neural networks have gained interest and are emerging as an alternative paradigm due to their outstanding performance against existing traditional methods \cite{cheng2020image,shukor2022video, reducinggap,xie2021enhanced, balcilar2022reducing}. There are two different paradigms: Variational Autoencoder (VAE) and Implicit neural representation (INR) based methods. VAE methods learn a transformation to project the data in the latent space and directly optimize the latents to minimize the rate (entropy)-distortion (mean squared error) loss function \cite{balle2018variational}, while INR have only recently emerged \cite{dupont2021coin} and their performance is below the former one. Our work focuses on this paradigm to improve INR based image compression method.

INR represents an image by over-fitting a continuous function (neural network) which takes as input coordinates and outputs the pixel color values \cite{sitzmann2020implicit}. This can be used for compression, since transmitting an image amounts to transmit the bit-stream of weights for the neural network that allows for image reconstruction. At decoding, the image can be reconstructed by extracting the weights from the bit-stream and evaluating the neural network on the coordinates. \cite{dupont2021coin} demonstrated the potential of INR for image compression by showing their ability to  outperform JPEG standard especially at low-bit rates. However, the main disadvantage is that they perform only naive compression by quantizing the weights into single (16 bits) precision and encoding time is high. Subsequently, \cite{dupont2022coin++,strumpler2021implicit} overcame this limitation by using meta-learned weights as initialization to decrease the encoding time and use a better quantization and arithmetic coding to improve compression performance. The idea of using INR for video compression is shown in \cite{chen2021nerv}. The existing work still suffers from several limitations: the decrease of performance due to the quantization error, non-optimal quantization, and inefficient entropy model. 
Furthermore compared to VAE based approaches, the performance of INR is not on par at the moment \cite{dupont2022coin++}, however the  decoding complexity and energy consumption of the VAE based approaches are several magnitudes higher compared to the traditional approaches, thus far from practical consideration. 

In this work, we first show the potential of INR based approaches from practical point of view by showing that they have a very lower decoding complexity compared to VAE based approaches.
Second, we propose several contributions to improve the performance of INR based image compression and we label our proposed method as RQAT-INR. Firstly, we propose a fixed-bit quantization with absolute maximum normalization scheme to quantize the weights of INR, secondly we propose a regularized quantization aware training model to revert the performance degradation due to the quantization, and finally we propose a border aware entropy model to efficiently model the weights distribution after the fixed-bit quantization. Experimental results demonstrate that our proposed method outperforms the existing methods \cite{dupont2021coin,dupont2022coin++} by $32-41\%$ net bit-rate savings.


\section{Background}\label{sec:method}
\begin{figure}
    \centering
    \includegraphics[scale=0.45]{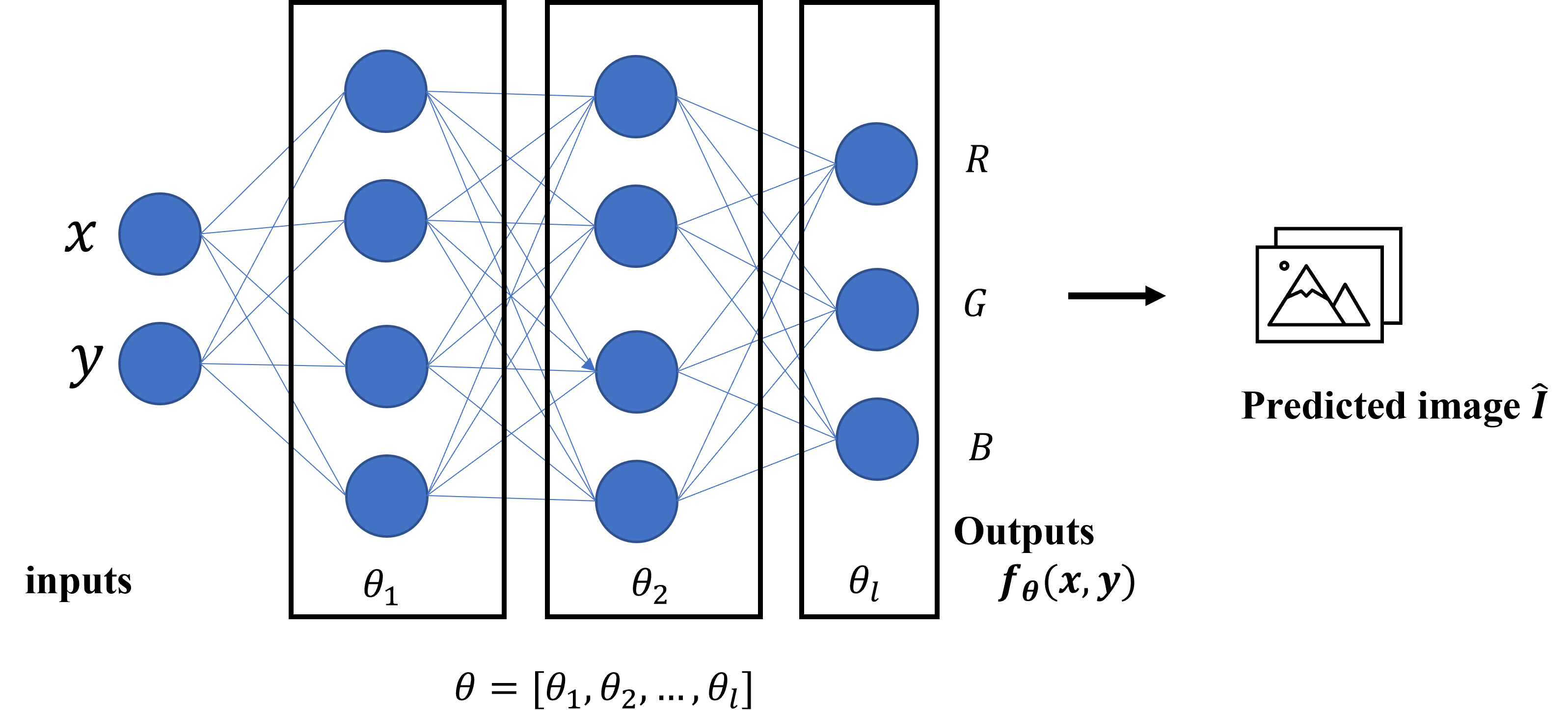}
    \caption{Implicit neural representation-based image compression method. The weights $\theta$ are encoded in the bit-stream and transmitted to the decoder side.}
    \label{fig:INR}
\end{figure}

Let $\mathbf{I} \in \mathbb{R}^{W \times H \times 3}$ be a color image, $x,y \in \mathbb{R}$ be the pixel coordinates in the normalized range $[-1, 1]$, $I(x,y)$ denotes the pixel values (RGB) at the coordinates $x,y$. The INR is a function $f_\theta$, parameterized by the neural network with weights $\mathbf{\theta}$ such that it maps the given coordinates to the pixel intensity values (RGB). In other words, $\forall x,y, f_\theta(x,y) \sim I(x,y)$. The weights $\mathbf{\theta}$ of the INR are obtained by over-fitting (minimizing) the following loss function
\begin{equation}\label{eq:inr_mse}
L=\frac{1}{W\times H} \sum_{x,y} d(I(x,y),f_\mathbf{\theta} (x,y)),
\end{equation}
where the sum is over all the pixels in the image $(W\times H)$, $d$ is any distortion metric which measures the discrepancy between the predicted (reconstructed) pixels by $f_\theta$ and the actual pixel values of the image $I$. The metric $d$ is preferably a differentiable distortion measure, such as mean squared error or perceptual metric such as LPIPS. Once the equation \eqref{eq:inr_mse} is optimized, the compressing an image $\mathbf{I}$ is equivalent to storing the values of the weights $\mathbf{\theta}$. Figure \ref{fig:INR} illustrates a simple implicit neural network (INR) based image compression system.

The INR is designed using a multi-layer perceptron (MLP) with sinusoidal activation functions\cite{sitzmann2020implicit} to capture the high frequency details in the underlying image \cite{dupont2021coin}.
For each image $I$, there is one specific INR $f_\theta$ which is overfitted to the given image I. The quality of the reconstructed image by $f_\theta$ depends on the size of the neural network, but here in the compression task we cannot choose large size neural network because it will increase bitlength, as the weights are used as the descriptions of the image, thus the number of weights are constrained  at the expense of the distortion. Compared to VAE based image compression, rate(R)-distortion(D) trade-off in INR based image compression is controlled with respect to the number of weights or size of the neural network. So, for different rates, the INR has a different neural network architecture with a different number of weights.

\section{Proposed method}\label{sec:pro}
Our proposed method consists of three major components: we first describe  the quantization scheme, then the entropy coding of the weights, and finally the regularized quantization aware training. 

\subsection{Absolute maximum normalized quantization}\label{subsec:quant}
Let $L$ be the number of layers in MLP, $\theta_w=[\theta_{w^1} \theta_{w^2},\dots,\theta_{w^L} ]$ be the collection of weights, and $\theta_b=[\theta_{b^1}, \theta_{b^2},\dots,\theta_{b^L}]$  be the collection bias of all the layers with full precision, and $\theta=[\theta_w,\theta_b ]$. Let q be the number of fixed bits used to quantize the weights (q=8, for 8-bit quantization) and $k=2^q/2-1$. Now fixed-bit quantization (q) of $\theta_w$ for the $l^{th}$ layer is performed as follows: 
\begin{itemize}
    \item 	First, the absolute maximum is computed over the weights
$w_{max}^l=\ \max{\left({|{\theta}}_{{w}^{l}}\middle|\right)}$.
	\item Second, the weights are normalized with respect to the $w_{max}^l$,
${\theta}_{{n}{w}^{l}}=\ \frac{{\theta}_{{w}^{l}}}{w_{max}^l}$.

\item 	Third, fixed-bit quantization are computed as
$Q\left(\theta_{w^l}\right)={\hat{{\theta}}}_{{w}^{l}}=round\left( \theta_{nw^l} k\right)$, where $round$ converts to the nearest integer.

\item Fourth, after quantization $Q\left({\theta}_{{w}^{l}}\right)$, the dequantization is performed as
$\tilde{\theta}_{w^l}$ = $Q^{-1}\left({\hat{{\theta}}}_{{w}^{l}}\right)= \frac{{\hat{{\theta}}}_{{w}^{l}}}{{k}}w_{max}^l$.
\end{itemize}

The above steps are repeated for the all the remaining layers $l=1,2,\ldots,\ L$, and the fixed-bit quantized weights of all the layers is denoted as ${\hat{{\theta}}}_{w}=\left[{\hat{{\theta}}}_{{w}^\mathbf{1}},\ldots,{\hat{{\theta}}}_{{w}^{L}}\right]$. In a similar manner, we also perform fixed-bit quantization of bias of all the layers and denote as  ${\hat{{\theta}}}_{b}=\left[{\hat{{\theta}}}_{{b}^\mathbf{1}},\ldots,{\hat{{\theta}}}_{{b}^{L}}\right]$. While decoding the quantized weights from the bit-stream, the absolute maximum value of weights and bias of the all the layers are required to perform dequantization. So these values (the absolute maximum value of weights and bias of the all the layers) are explicitly encoded in the bit-stream using a 16 bits representation. 
Thus, we spend $2\times L\times16$ extra bits in addition to the network weights.


\subsection{Border-aware entropy model}\label{subsec:ent}

Instead of storing quantization weights using $q$-bits, we use entropy coding to gain additional compression efficiency. For this, we take advantage of the weight distribution shape, and model the $q$-bit quantized weights $\hat{{\theta}}=[{\hat{{\theta}}}_{w},{\hat{{\theta}}}_{b}]$ in section \ref{subsec:quant} to follow explicit univariate probability distribution. However, directly encoding the data with univariate probability distribution might not predict well probabilities of (extreme values) in the $q$-bit quantized weights. For this, we propose an entropy model which is aware of its border (extreme) values. 

We propose to use fixed probability for the border values $-k$ and $k$ (i.e -127 and +127, for 8-bit quantization) and gaussian distribution for the rest of the symbols. It is because in every layer, there is at least one symbol whose value is the absolute maximum (either positive or negative). This symbol can be either $-k$ or $+k$ and their probabilities cannot fit any gaussian distribution well. Since there are $\left|\hat{{\theta}}\right|=\left|{\hat{{\theta}}}_{w}\right|+\left|{\hat{{\theta}}}_{b}\right|$ number of parameters to be encoded and at least L out of $\left|{\hat{{\theta}}}_{w}\right|$ weights and L out of $\left|{\hat{{\theta}}}_{b}\right|$ biases should be quantized either $-k$ or $+k$  with the same probability, we can write the probability of parameters being $-k$ or $+k$  with $p\left(-k\right)=p\left(k\right)=L/\left|\hat{{\theta}}\right|$. We assume the rest of the symbols follow the truncated gaussian distribution with the support of [-(k-1), (k-1)] and total probability of $1-2L/\vert\hat{{\theta}}\vert$. The parameters of the gaussian distribution can be calculated by encoded symbols’ statistics whose values are not $-k$ or $+k$ . Thus, if we define parameters to be encoded whose value is not $-k$ or $+k$  by $\bar{{\theta}}=\left[\theta\in\hat{{\theta}}\ \right|(k-1)\geq\theta\geq-(k-1)]$, we can show the parameters of the Gaussian distribution’s mean $\mu=E(\bar{\theta})$ and variance $\sigma^2=E(\bar{\theta}^2)-E(\bar{\theta})^2$ estimated from $\bar{{\theta}}$. Thus, the probabilities of each symbol can be shown by followings, if $N\left(.;\mu,\sigma^2\right)$ is the Gaussian distribution with given parameters,
\begin{equation}\label{eq:prob}
p\left(x\right)=
\begin{cases}
\left(1-\frac{2L}{|\hat{\theta}|}\right).\frac{N(x;\mu,\sigma^2)}{\sum_{u=-(k-1)}^{(k-1)} N(u;\mu,\sigma^2)}, & \text{if} \ (k-1) \geq x \geq -(k-1) \\

L/|\hat{\theta}|,  & \text{if} \   x =k  \ \text{or} \ x=-k \\
0, & \text{else}                                                                      
\end{cases}
\end{equation}

The rate (expected bit-length) of the $\hat{{\theta}}$ can be computed as 
\begin{equation}
R=\ -\sum_{i}{\log_2{p(}{\hat{{\theta}}}_i)}
\end{equation}

The quantized weights $\hat{{\theta}}$ are encoded using the CDF from \eqref{eq:prob} in the bit-stream using entropy coding \cite{duda2009asymmetric}. We also need to encode the mean ($\mu$ ) and variance ($\sigma^2$) of the distribution in the bit-stream using $16$ bits floating point precision for each. 

\subsection{Regularized quantization aware training}
The steps in section \ref{subsec:quant} and \ref{subsec:ent} are sufficient to encode the quantized weights in the bit-stream, however there will be a large performance degradation depending on the bits ($q$) used for quantization. Quantization aware training (QAT) \cite{nagel2021white} can be used to revert this, but the degradation still exists. We overcome this by proposing regularized quantization aware training which adds an regularization term on the loss function \eqref{eq:inr_mse}. Instead of just using distortion between quantized model’s prediction and original image with $d\left(I\left(x,y\right),f_{\hat{{\theta}}}\left(x,y\right)\right)$, we also use distortion between quantized model’s prediction and full-precision model prediction as a regulation term in the loss function with a hyperparameter $\lambda$. Thus, during the training, we minimize following loss function
\begin{equation}\label{eq:reg_loss}
L=\frac{1}{MN}\sum_{x,y}{d\left(I\left(x,y\right),f_{\hat{{\theta}}}(x,y\right))}+\lambda d\left(f_{{\theta}^\ast}\left(x,y\right),f_{\hat{{\theta}}}\left(x,y\right)\right),
\end{equation}
where $f_{{\theta}^\ast}$ is the optimized network with full precision weights (32-bit floating point), and it is fixed through out the training. The choice of the hyperparameter $\lambda$ can be chosen for the whole dataset, or it can be tuned according to the specific image. The regularization term in \eqref{eq:reg_loss} can also be viewed as the knowledge distillation, where the knowledge is distilled from the full precision optimized network to the quantized network. 

During the backward pass, as the nature of quantization is non-differentiable, the gradients are approximated using straight-through-estimator (STE), and weights are updated with stochastic gradient descent. Following, the weights and bias are quantized to $q$-bits. This process is repeated for each optimization step until convergence or a certain number of iterations. The advantage of using our proposed regularized quantization aware training is twofold: convergence of training is faster and it decreases the impact of the quantization error (lower mse).

Once the regularized quantization aware training is completed, instead of writing quantized weights $\hat{{\theta}}$ using $q$-bits, we entropy code the quantized weights $\hat{{\theta}}$ as mentioned in section \ref{subsec:ent}.

For an given image, the information encoded in bitstreams contains
\begin{itemize}
    \item 	$q$-bit quantized weights $\hat{{\theta}}$, encoded by an range encoder,
    \item  Absolute maximum values of weights ${\theta}_{w}$ and bias ${\theta}_{b}$, encoded using $2\times L\times16$ bits, where $L$ is the number of layers,
	\item Mean and variance of the normal distribution, encoded using $2\times16$ bits
\end{itemize}

\subsection{Decoding}
\begin{figure}
    \centering
    \includegraphics[scale=0.6]{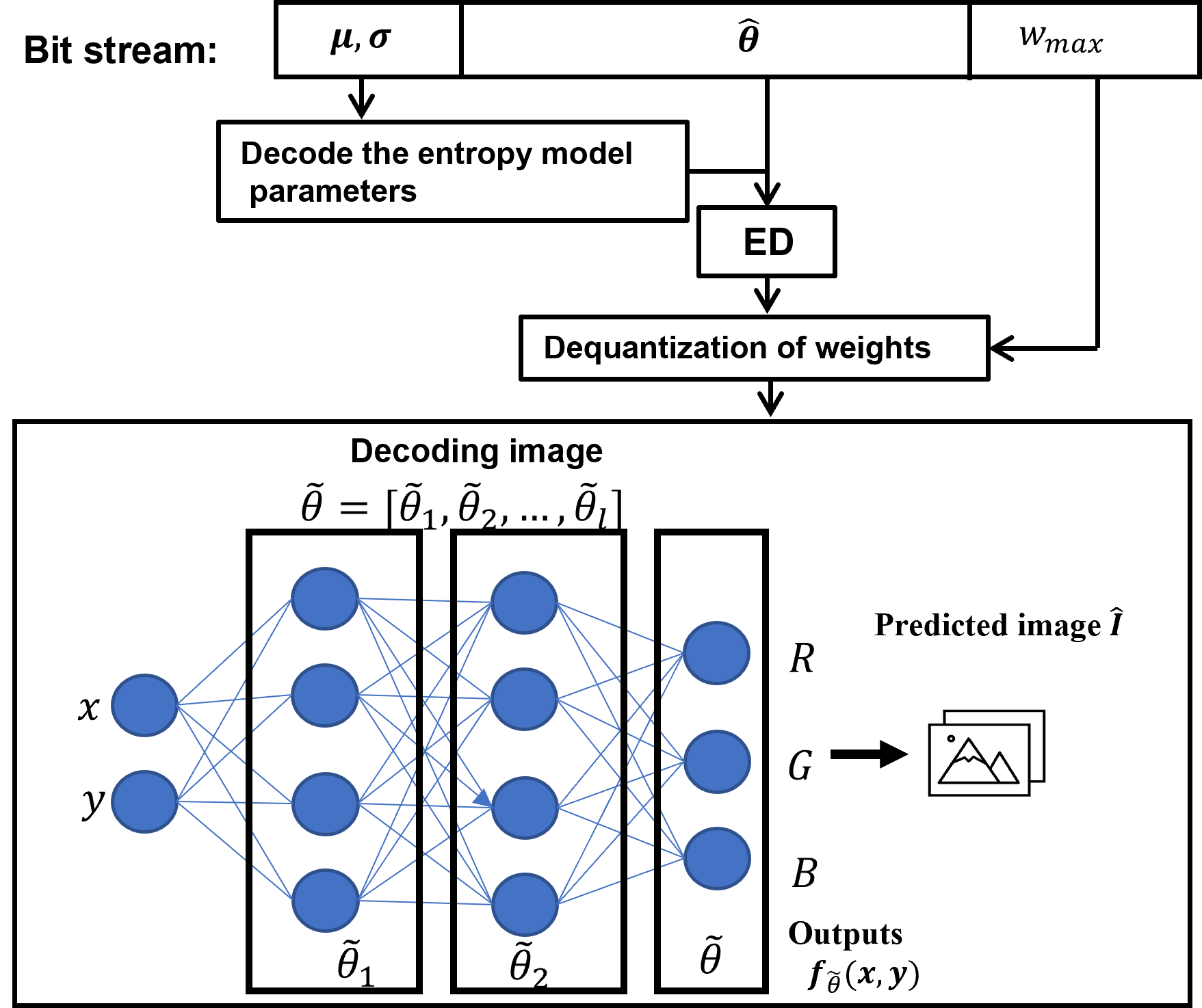}
    \caption{Decoding procedure from the bit-stream to reconstruct the image. ED denotes entropic decoder.}
    \label{fig:decoding}
\end{figure}

Decoding is illustrated in Figure \ref{fig:decoding}. The mean and variance are decoded from the bitstream to decode the quantized weights $\hat{{\theta}}$. Then, the absolute maximum values of weights and bias are decoded. Finally, the decoded quantized weights $\hat{{\theta}}$ are de-quantized using inverse quantization to obtain $\tilde{\theta}$, that is to scale the weights prior to quantization. Finally, decoding of the image is just a forward pass of the corresponding network on the pixel coordinates.

\section{Experimental Results}
In order to demonstrate the effectiveness of our proposed method, RQAT-INR, we compare our method with the competitors: COIN \cite{dupont2021coin}, and COIN++\cite{dupont2022coin++} on the Kodak dataset \cite{eastman_kodak_kodak_nodate} over different bit-rates. We have also included JPEG, and JPEG2000 only for the reference with traditional methods and we only base our comparisons with INR-competitors. The PSNR metric on RBG and bits per pixel (BPP) are used as the distortion and rate measure.  We use the same architecture (number of hidden nodes, and layers) for MLP as in the competitors \cite{dupont2021coin, dupont2022coin++} for different bit-rates, as well as the same optimization procedures (Adam with lr= $2e-4$). For our proposed method, the weights are initialized with optimized $32$-bit precision weights of baseline method COIN, and trained for $15K$ iterations. It can be initialized with random weights, but optimized 32-bit precision weights resulted in better performance and convergence speed. The hyperparameters $\lambda$ is varied over $\{0.005, 0.05, 0.01, 0.1\}$. Regarding the  quantization bit ($q$) resolution, we used $8$-bits for the low bpp and $10$-bits for high bit rate regime (last two points in the RD curve).     

\begin{figure}[!htbp]
  \centering
    \includegraphics[width=.49\textwidth]{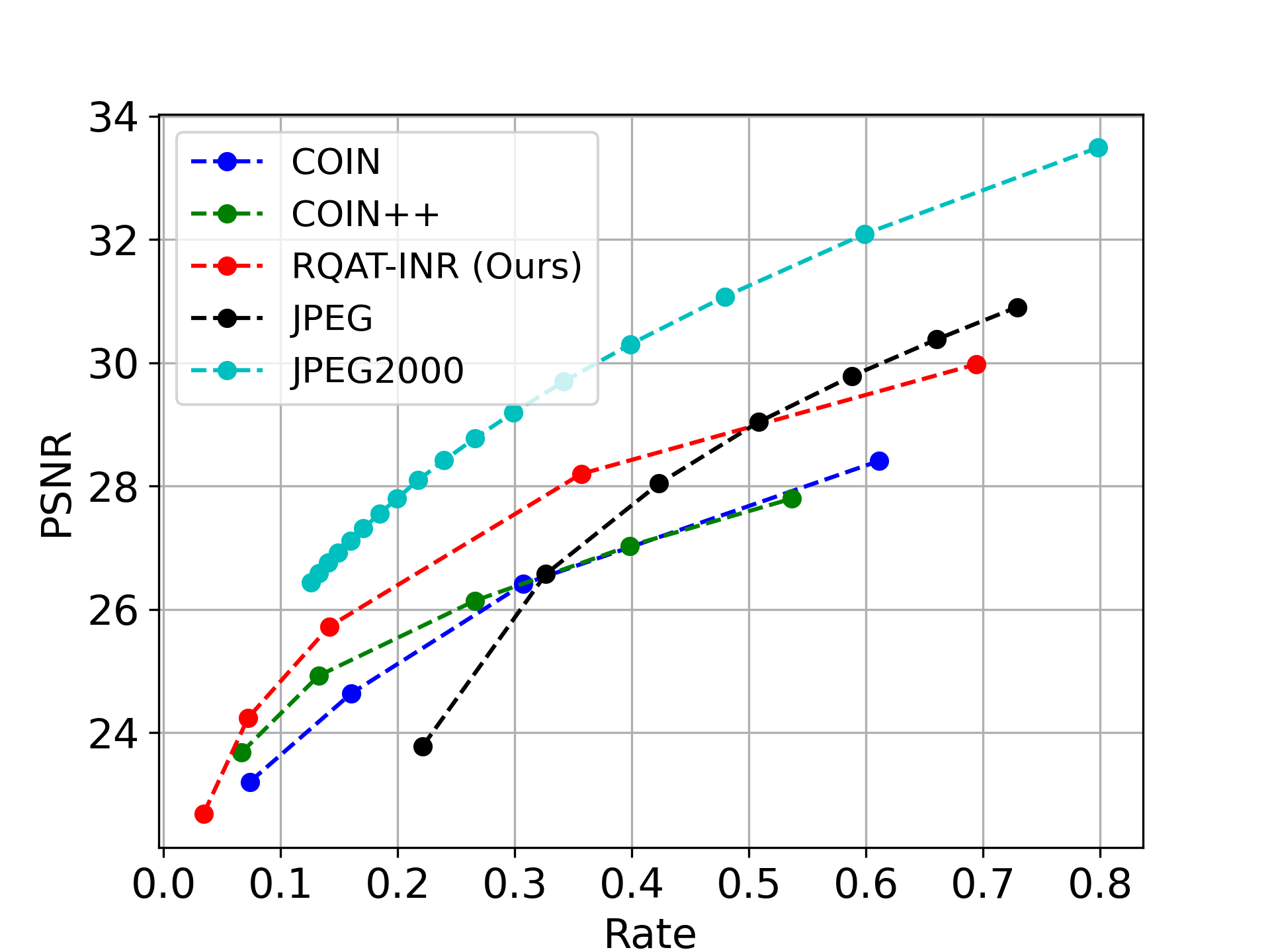}
    \includegraphics[width=.49\textwidth]{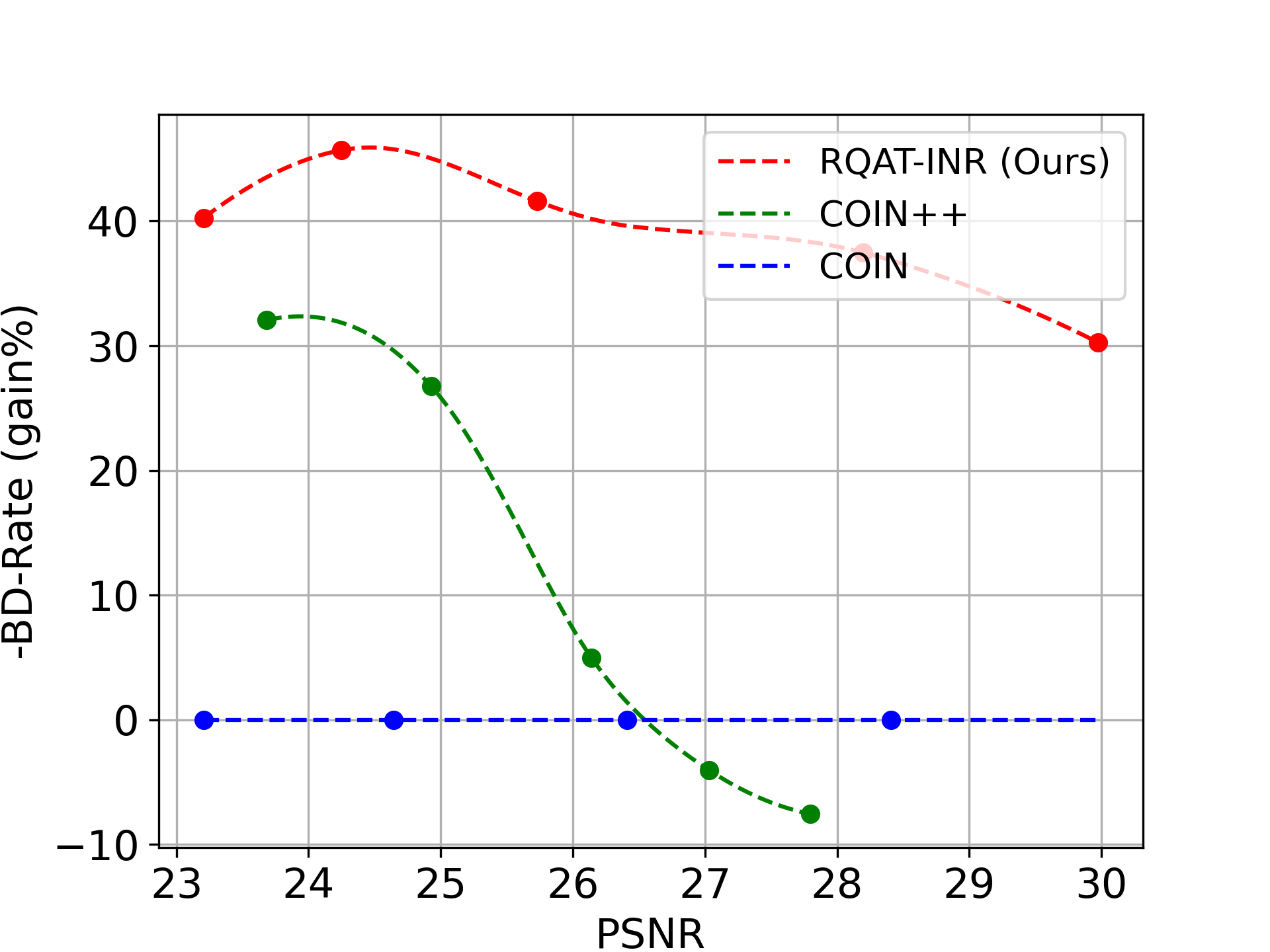}
  \caption{Left: Rate distortion curve of our proposed method and competitors averaged over all the images in the kodak data set (JPEG and JPEG2000 are included for the reference). Our method significantly outperforms the competitors across the bit-rate regimes. Right: Bjontegard BD rate gain of our method compared to the COIN and COIN++ on the kodak dataset. Our method has 41\% and 32\% net bitrate savings over COIN and COIN++.}
\label{fig:coinrd}
\medskip
\end{figure}

The rate distortion curve of our proposed method and competitors computed on the Kodak dataset is displayed in fig \ref{fig:coinrd}a, and can be observed that our method significantly outperforms the competitors across the bit-rate regimes. The competitor COIN++ only performed better than the baseline (COIN) in the low bpp, and in high bpp it is similar or slightly lower.   To quantify the bit-rate gain in \%, we computed Bjontegard BD rate gain \cite{bjontegaard2001calculation} and it demonstrated that our method has about 41\% and 32\% gain over the COIN and COIN++ (see figure \ref{fig:coinrd}b) respectively. Our method resulted in higher gain over COIN++ in the high bit-rate regime. To show the advantage of our regularization term, we also performed experiments without regularization term, and we observed about 7\% gain over the quantization aware training without regularization term. Thus, highlighting the potential of our method to bridge the performance gap due to the quantization error of the weights.

\begin{figure}[!htbp]
  \centering
    \includegraphics[width=.49\textwidth]{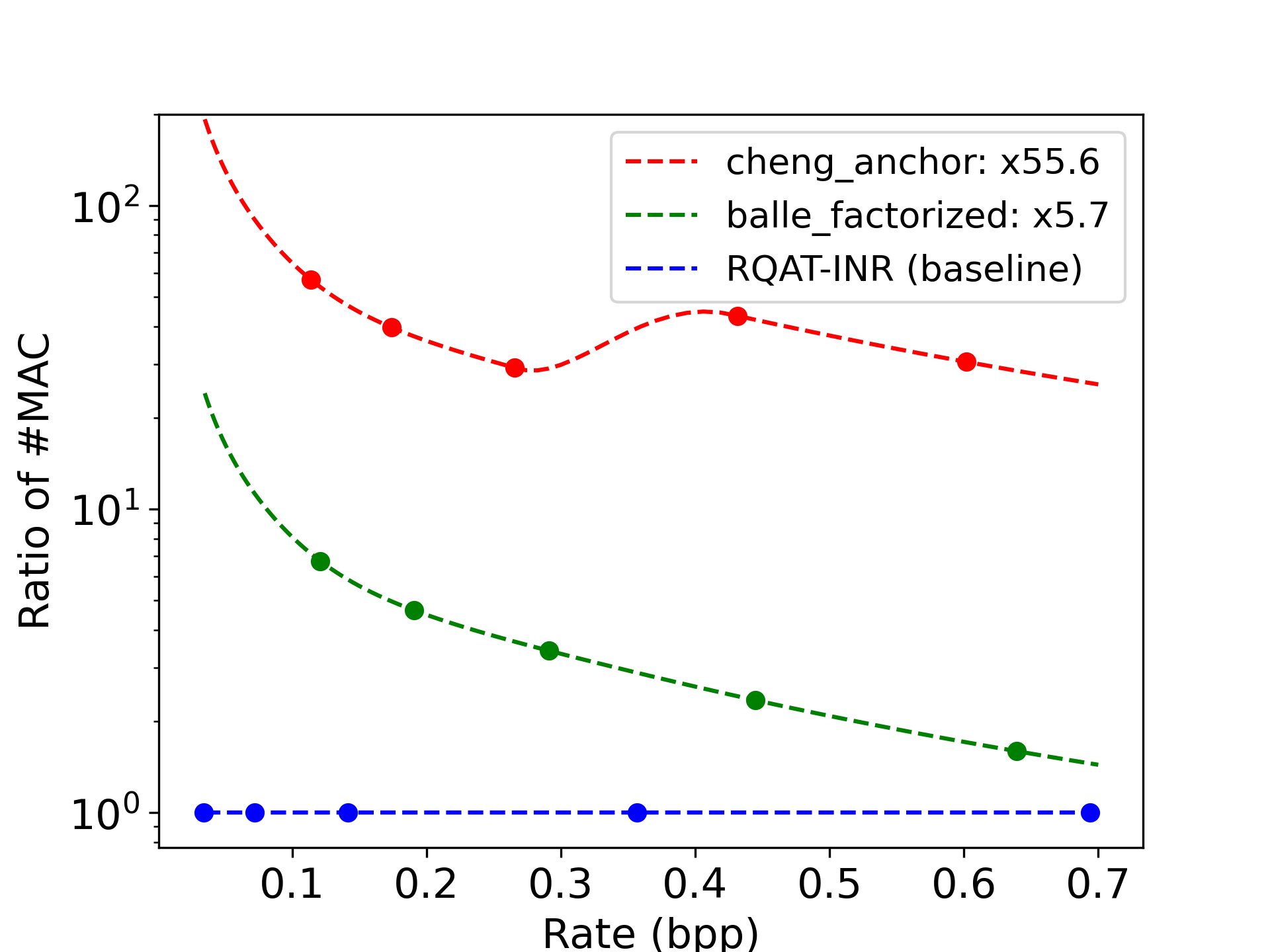}
    \includegraphics[width=.49\textwidth]{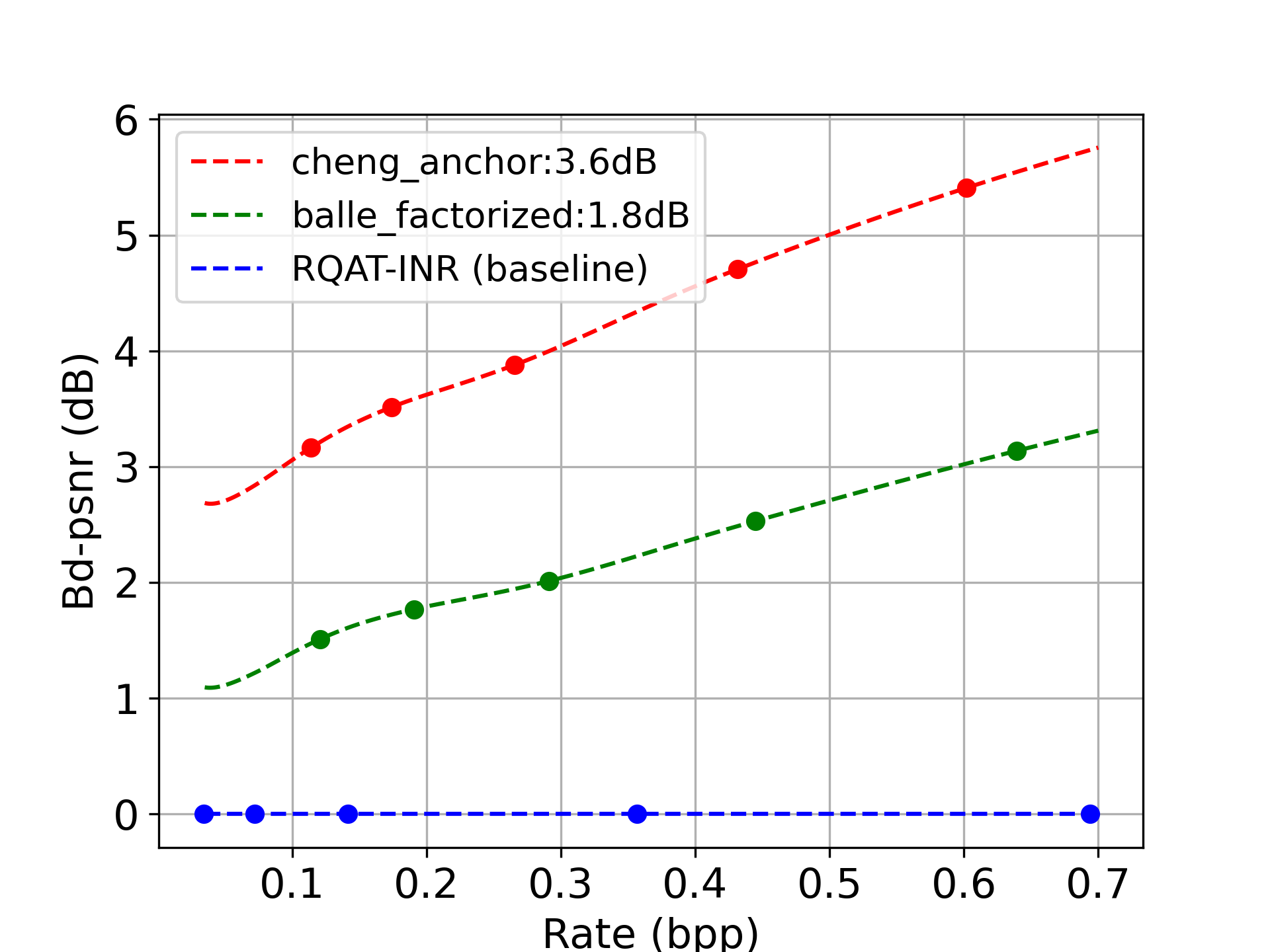}
  \caption{a) VAE based neural codecs' relative number of MAC per pixel for the same rate wrt our proposal.  \textbf{cheng\_anchor} needs almost 55 times MAC than our INR while \textbf{balle\_factorized} needs almost 6 times for the same rate in average between 0.04-0.7 bpp.  b) VAE based neural codecs' Bd-psnr for the same rate wrt our proposal.  \textbf{cheng\_anchor} has 3.6dB better PSNR reconstruction than our INR while \textbf{balle\_factorized} has 1.8dB better PSNR for the same rate in average between 0.04-0.7 bpp.}
\label{fig:coin_flops}
\medskip
\end{figure}


Now, we show the decoding complexity of the VAE and INR based neural codecs. We used \textbf{cheng\_anchor}, the best performing method in \cite{cheng2020image} in terms of RD performance and \textbf{balle\_factorized}, the least decoder complexity model in \cite{balle2016end} from compressAI library \cite{compressai} for VAE based method, and our \textbf{RQAT-INR} codec that includes defined improvements over COIN. We used two different tools to count the number of floating point operations during decoding. The first one is PAPI \cite{papi} which counts the experimental floating point operations (flops). Since this tool reports the number of floating point operations on single cpu and single thread in hardware level, we run the test on single core of Intel(R) Core(TM) i7-8850H CPU @ 2.60GHz with preventing multi-thread operations. In order to do crosscheck, we also reported the theoretical number of multiplication accumulation operator (MAC) using FVCORE \cite{fvcore}. This tool basically counts every multiply-add operation as one operation. The complexity test results alongside with average rate and PSNR of Kodak dataset for provided all 5 quality models are given in Table \ref{tab:complexity}. As it can be seen that models average rate's and distortion's are not comparable, but at least the number of operations give some idea about models complexity in Table \ref{tab:complexity}. In average, the number of MAC is excepted to be half of the flops number. Because one MAC operation includes two operations which are one multiplication that follows by one addition operation. The differences can be explained by the shortcomings of the used tools to count them. For instance, PAPI counts the operation in hardware level and the results may be different on different hardware. Also, FVCORE does not count operations that are performed out of common layers. In addition, some common layers are still missing in FVCORE's results. These explains some inconsistency with flops and MAC results in Table \ref{tab:complexity}.    

In order to do comparison between models, we show the relative MAC per pixel number for the same rate in Figure \ref{fig:coin_flops}a. It can be seen that our RQAT-INR codec's decoder is 55 times less complex than most performing codec \textbf{cheng\_anchor} while almost 6 times less complex than the least complexity VAE based model \textbf{balle\_factorized} for the same rate. However these numbers cannot tell the entire truth without Figure \ref{fig:coin_flops}b which explains our loss in terms of PSNR for having this low complexity for the same rate. It can be seen that our RQAT-INR's PSNR result is 1.8dB lower than \textbf{balle\_factorized} and 3.6dB lower than \textbf{cheng\_anchor} in average. Even though there is big differences in the reconstruction quality, INR based model still might be preferable in some specific cases because of lower complexity.

\begin{table}[h] 
\centering
\caption{Rate (bpp), Psnr (dB), number of kflops per pixel and number of kMACS per pixel average for Kodak test set and provided first 5 quality model.}\label{tab:complexity}
\begin{tabular}{lcccc}
\toprule
{} & Rate (bpp) & PSNR (dB) & kflops/pxl & kMAC/pxl \\
\midrule
RQAT-INR & 0.2597 & 26.16 &  25.91 & 11.26  \\
balle\_factorized  & 0.3374 & 29.79 & 124.23 & 61.49  \\
cheng\_anchor & 0.3162 &  31.56 & 840.52 & 546.46 \\
\bottomrule
\end{tabular}
\end{table}

\section{Conclusion}
In this paper, we demonstrated that INR based compression approach has a more practical advantage than the VAE based approaches due to it's lower decoding complexity. Further we proposed to improve the compression rate of the implicit neural representation based image compression by proposing regularized quantization aware training and border aware entropy model. Our method brings about 32-41\% bit-rate gain compared to the existing methods. However, this improvement is not enough for now to be competitive to the state of the art models especially on high rate regimes. As it can be seen in Figure \ref{fig:coin_flops}b, INR's PSNR loss increases by rate. It can be explainable by the fact that selected MLP architecture is not the optimal architecture for image approximation under entropy constraints. At lower rates, since the number of parameters is lower, the possible architecture of the network is also limited. Thus, selected MLP may not be far away from the optimal architecture. However, in high rate regime, the possible architecture increases exponentially by the increment of number of parameter of the INR network. Thus, looking for learning the best universal architecture and/or image adaptive architecture would be some way of get rid of this problem and we propose to address this issue in future work. In addition, our quantization is for encoding the weights into bitstream but not decreasing the complexity of the model. Our decoder still performs single precision floating point operations on dequantized parameters. It is in our future work to implement INR's decoder network with integer network in order to decrease the complexity more.



\Section{References}
\bibliographystyle{IEEEbib}
\bibliography{refs}

\end{document}